\documentclass[twoside,11pt]{article}

\usepackage{jmlr2e}
\usepackage{hyperref}
\usepackage{amsmath}
\usepackage[utf8]{inputenc}
\usepackage[english]{babel}
\usepackage{multirow}
\usepackage{hhline}
\usepackage{graphicx}
\firstpageno{1}

\begin{document}

\title{Recent Trends in Named Entity Recognition (NER)}

\author{\name Arya Roy \email aryar@alumni.cmu.edu \\
       \addr Carnegie Mellon University\\
       Pittsburgh, PA 15213, USA
       }

% \editor{Leslie Pack Kaelbling}

\maketitle

\begin{abstract}%   <- trailing '%' for backward compatibility of .sty file
The availability of large amounts of computer-readable textual data and hardware that can process the data has shifted the focus of knowledge projects towards deep learning architecture. Natural Language Processing, particularly the task of Named Entity Recognition is no exception. The bulk of the learning methods that have produced state-of-the-art results have changed the deep learning model, the training method used, the training data itself or the encoding of the output of the NER system. In this paper, we review significant learning methods that have been employed for NER in the recent past and how they came about from the linear learning methods of the past. We also cover the progress of related tasks that are upstream or downstream to NER eg. sequence tagging, entity linking etc. wherever the processes in question have also improved NER results.
\end{abstract}

\begin{keywords}
  named entity recognition, sequence tagging, deep learning, conditional random fields, bidirectional long short term memory, recurrent neural network, word embedding 
\end{keywords}

\section{Introduction}
% ~\citep{pearl:88}

‘Named Entity Recognition’ refers to the Natural Language Processing task of identifying important objects (eg. person, organization, location) from text. Henceforth we will use NER to refer to Named Entity Recognition and Classification. NER belongs to a general class of problems in NLP called sequence tagging ~\citep{erdogan:10}. Sequence Tagging NLP tasks other than NER are Part of Speech (POS) tagging and chunking. We present a survey of the latest research from the NER field. But a lot of the mentioned research pertains to sequence labelling tasks other than NER as well. Hence we consider the mentioned research as a comprehensive survey of sequence tagging with a focus on NER. Also, we show research 2006 onwards wherever possible since there has been a shift from linear or log-linear methods ~\citep{nadeau:07} to non-linear methods in the past decade. To the best of our knowledge, NER pre-processing, training and evaluation methods have not been surveyed extensively 2006 onwards.

The first section of the survey briefly covers the evolution of the deep learning framework from linear techniques ie. the advantages and drawbacks of each with respect to training, and performance in different domains, languages and applications. We cover around 150 well-cited English language papers from major conferences and journals. We do not claim this review to be exhaustive or representative of all the research in all languages, but we believe it gives a good estimate of the major research findings in the field.

Sections 2, 3 and 4  detail the computational techniques that have produced state-of-the-art results at various stages of the neural network model in order. In particular, Sections 2 covers feature engineering at the input i.e., syntactical, morphological and contextual features. Moreover, Section 2 shows the impact of different distributed representation models of the text on the NER task. There is also a discussion of the structured and unstructured data used  and additional preprocessing steps that have shown state-of-the-art NER results. Section 3 discusses the model architecture used in NER including convolutional, recurrent and recursive neural networks. Section 3 also shows the optimisation method, output tag structure and evaluation methods and their impact on the performance of particular techniques. Section 4 summarizes the performance of the discussed techniques on standard datasets used to benchmark NER performance.

\section{A Brief History}

Researchers have produced a diverse set of techniques to solve the NER task over close to 30 years. ~\citep{rau:91} solved part of the NER task as we know it at present. However there is previous research in Information Extraction that includes Named Entity Recognition in a more constrained manner than its present form ~\citep{besemer:87,dejong:79,dyer:86,grishman:14,hobbs:86,lytinen86,rau:87, young:85}. The initial NER research comprised of hand-crafted rules-based linear models that were overfitted to a very specific structured text corpus eg. military message sets,  naval operation reports, M\&A news ~\citep{jacobs:93} etc. The need for standardisation led to MUC-6 ~\citep{grishman:96}, HUB-4 ~\citep{chinchor:98}, MUC-7 and MET-2 ~\citep{chinchor:97}, IREX `\citep{sekine:00}, CONLL ~\citep{sang:03}, ACE ~\citep{doddington:04} and HAREM ~\citep{santos:06}. The Language Resources and Evaluation Conference (LREC)1 has also been staging workshops and main conference tracks on the topic since 2000.  Supervised learning techniques trained on a large annotated corpus produced state-of-the-art results for NER. The prominent supervised learning methods include Hidden Markov Models (HMM) ~\citep{bikel:97}, Decision Trees ~\citep{sekine:98}, Maximum Entropy Models (ME) ~\citep{borthwick:98}, Support Vector Machines (SVM) ~\citep{asahara:03}, Conditional Random Fields (CRF) ~\citep{lafferty:01,mccallum;03}. CRF in particular is one of the most effective NER algorithms. The need in NER to train the probability of the output tags using many leading and lagging non-local sequences makes a discriminative model like CRF more suitable compared to generative models like HMM and stochastic grammar. Although ME models relax the strong independence assumptions that the generative models make, they suffer from a weakness called the label bias problem, wherein the model is biased towards states with few outgoing transitions. CRF solves the problem by jointly considering the weights of different features in all the states instead of normalising the transition probabilities at the state level. ~\citep{sarawagi:04} improved on the existing CRF model by formulating the semi-CRF model that assigns labels to subsequences instead of individual entities without any significant additional computational complexity. `\citep{passos:14} used their lexicon-infused skip-gram model on public data to learn high-quality phrase vectors to plug into a log-linear CRF system. Joint models of entity analysis ( multitask models) have shown better results on individual tasks compared to models optimised solely for NER. ~\citep{durrett;14} developed a structured CRF model that improves the performance of NER ( and also that of other entity analysis tasks) by training for 3 tasks eg. coreference resolution (within-document clustering), named entity recognition (coarse semantic typing), and entity linking (matching to Wikipedia en- tities). ~\citep{luo:15} proposed the JERL ( Joint Entity Recognition and Linking) model ie. semi-CRF model extended to capture the dependency between NER and entity linking. 
Supervised learning methods hit a roadblock because there is a limit to the structured text available for learning discriminative features. This led to Semi-supervised learning methods that utilise the exponentially growing volume of unstructured text ie. webpages to derive contextual information in an unsupervised manner from the seed annotated corpus ie. bootstrapping ~\citep{nadeau;06, brin:98,collins:99, yangarber:02, riloff:99, cucchiarelli:01, pasca2006}. ~\citep{suzuki:11} proposed an unsupervised method to create resourceful condensed feature representations from large scale unlabelled data to efficiently train supervised NER systems while still maintaining state-of-the-art results derived from existing higher dimensional semi-supervised solutions.

\begin{figure}[h]
\includegraphics[scale=0.5]{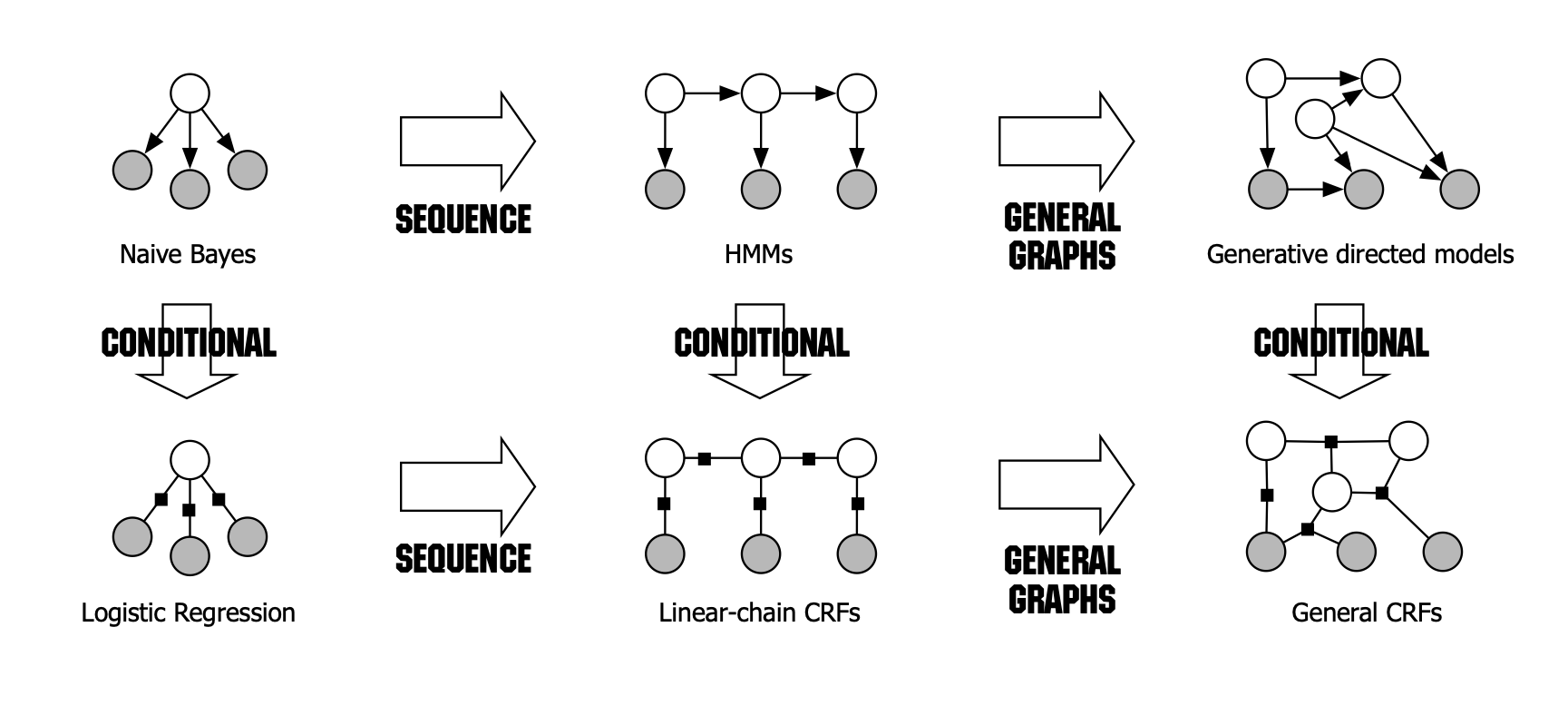}
\caption{Diagram of the relationship between naive Bayes, logistic regression, HMMs, linear- chain CRFs, generative models, and general CRFs ~\citep{sutton:06}}
\centering
\end{figure}
Unsupervised learning approach emerged mainly as the means to generate additional features from the context of input words to use in conjunction with other NER methods ~\citep{evans:03,cimiano:05, shinyama:04, etzioni:05}. More recently, ~\citep{lin:09} got state-of-the-art results without  using a gazetteer by instead performing k-means clustering over a private database of search engine query logs to use the cluster features to train their CRF model. An exception to the industry application of unsupervised learning technique is the clustering method using lexical resources eg. Wordnet ~\citep{alfonseca:02} in order assign named entity type from Wordnet itself (location>country, act>person etc.).
There have been many papers on employing neural networks to do NER in spite of the widespread adoption of CRF for the purpose. ~\citep{ratinov:09} used non-local features, a gazetteer extracted from Wikipedia, and Brown-cluster-like word representations in  a perceptron model \citep{freund:99}. Since multilayer feedforward neural networks are universal approximators, such a neural network could also potentially solve the NER task. ~\citep{petasis:00} used a feed-forward neural network with one hidden layer on NER and achieved state-of-the-art results on the MUC6 dataset. They used gazetteer and POS tags as input features. ~\citep{mikolov:13a} introduced the practise of using the skip-gram model or the continuous bag of words (CBOW) model to create a well-behaving and concise vector representation of words ie. word embeddings, upstream to the neural network model. ~\citep{collobert:08} demonstrated the use of pre-trained word embeddings to train a deep neural network to achieve state-of-the-art results in multiple NLP tasks including NER without any additional syntactic features. We discuss word embeddings in further detail in section 3. ~\citep{collobert:11} presented SENNA, which employs a deep FFNN and word embeddings to achieve near state of the art results on NER among other NLP sequence labelling tasks. ~\citep{santos:06} presented their CharWNN network, which extends the neural network of ~\citep{collobert:11} with character level CNNs, and they reported improved performance on Spanish and Portuguese NER. ~\citep{hammerton:03} attempted NER with a single- direction LSTM network and a combination of word vectors trained using self-organizing maps for lexical representation and principal component analysis for the orthogonal representation of POS and chunk tags. ~\citep{huang:15} used a BLSTM with extensive feature engineering instead of character embeddings for the POS-tagging, chunking, and NER tasks. ~\citep{chiu:15} used a LSTM-CNN model to detect character level and word level features without additional feature engineering for NER. Their model is similar to the RNN-CNNs model introduced by ~\citep{labeau:15} for German POS tagging but uses the advanced LSTM component compared to the RNN. ~\citep{lample:16} showed a BLSTM-CRF model and a stack LSTM model (s-LSTM) ~\citep{dyer:15} to perform the NER task by modeling the output tag dependency using a simple CRF model or a transition model to create and label chunks of input text. Like ~\citep{lample:16}, ~\citep{yang:16} also used character level embeddings similar to ~\citep{ling:15}. ~\citep{yang:16} also used the deep hierarchical RNN ~\citep{cho:14} for sequence tagging.

\section{Features and Data}

In this section we look at the input to the NER model. The primary input is the training data. In order to benchmark NER performance, researchers run their model on the CoNLL-2002 and CoNLL- 2003 datasets ~\citep{sang:02, sang:03} that contain independent named entity labels for English, Spanish, German and Dutch. All datasets contain four different types of named entities: locations, persons, organizations, and miscellaneous entities that do not belong in any of the three previous categories. The miscellaneous category is very diverse as it includes adjectives, like Indian, as well as  events, like 1000 Lakes Rally. In the CoNLL-2003 dataset, Named entity tagging of English and German training, development, and test data, was done by hand at the University of Antwerp.
Almost all the research validates the hypothesis that NER systems perform better when there is external data. The external knowledge that NER systems use are gazeteers and unlabeled text. 

\subsection{Unlabelled Text}

Recent successful semi-supervised systems ~\citep{ando:05, suzuki:08} have illustrated that unlabeled text can be used to improve the performance of NER systems. ~\citep{ratinov:09} uses the implementation of the word class models ~\citep{brown:92} in ~\citep{liang:05} to obtain word clusters from the Reuters 1996 dataset, a superset of the CoNLL03 NER dataset. ~\citep{ratinov:09} produced unique bit strings for each word based on its path from the root in the binary tree that the word class algorithm produces. They multiplied the input features by 4 by path representations of length 4,6,10 and 20 for every word.

~\citep{qi:09} proposed an iterative Word Class Distribution Learning framework and applied it to a sampled set of Wikipedia webpages. In contrast to self-training (eg. bootstrapping) or co-training methods, the WCDL does not add self-assigned labels that might be subject to learning bias if the model introduces incorrectly labelled examples to the corpus. The WCDL iteratively re-trains a base classifier to build a class label distribution for each word by normalising the predicted labels of the unlabelled corpus. The word class distribution becomes a feature of the base classifier ( semi-supervised or supervised NER system) instead of adding many self-annotations, thus making the WCDL highly scalable.  
    
~\subsection{Gazeteers}

An important question in NER research is to address the coverage and the disambiguation of input words using a gazetteer ie. lists of words. There is ample evidence of the improvement of NER performance with the use of high-quality and high-coverage gazetteers ~\citep{cohen:04, torisawa:07, toral:06, florian:03}.
Wikipedia is a great source to construct gazetteers for NER for several reasons. (1) It is manually updated regularly hence there is less chance of missing new information. (2) It maps several variations of spelling or meaning to a relevant entry. For example, ‘Manchester’ not only refers to the English city of Manchester but also to the football club by the similar name. (3) Wikipedia entries are manually mapped to categories. For example, the entry about the city of Manchester is tagged as a city while Manchester City F.C. is tagged as a football association.  

Table 4 summarizes the results of the techniques for injecting external knowledge. Although the additional features were from the superset of the CoNLL03 dataset and the gazetteers were from Wikipedia, the extra features proved to be useful on all datasets.
To make the clusters more relevant to this domain, we adopted the following strategy: (1) Construct the feature vectors for 20 million phrases using the web data (2) Run K-Means clustering on the phrases that appeared in the CoNLL training data to obtain K centroids (3) Assign each of the 20 million phrases to the nearest centroid in the previous step.

\begin{equation}
    \begin{split}
        [y_s],[y_{s-1:s}],{[y_s,w_u]}_{u=s-1}^{s+1}, {[y_{s-1:s},w_u]}_{u=s-1}^{s+1} \\
    {[y_s,sfx3_u]}_{u=s-1}^{s+1}, {[y_{s-1:s},sfx3_u]}_{u=s-1}^{s+1} \\
  \{[y_s,wtp^t_u]_{u=s-1}^{s+1}\}^4_{t=1}, \{{[y_{s-1:s},wtp^t_u]}^{s+1}_{u=s-1}\}^4_{t=1} \\
   {[y_s,w_{u-1:u}]}_{u=s}^{s+1}, {[y_{s-1:s},w_{u-1:u}]}_{u=s}^{s+1} \\
   \{{[y_s,wtp^t_{u-1:u}]}_{u=s}^{s+1}\}^3_{t=1}, \{{[y_{s-1:s},wtp^t_{u-1:u}]}^{s+1}_{u=s}\}^3_{t=1}
    \end{split}
\end{equation}
  
Here, s denotes a position in the input sequence;
$y_s$ is a label that indicates whether the token at
position s is a named entity as well as its type; $w_u$
is the word at position u; sfx3 is a word’s three-letter suffix. ${wtp^t}^4_{t=1}$ are indicators of different word types; t=1 denotes punctuation, 2 denotes whether a word is upper cap, lower cap or all caps, 3 denotes a number, 4 denotes that the word has hyphen with different caps before and after the token. Including word unigram features ( to capture abbreviated or partial form of complete entities), ~\citep{lin:09} had 48 features. 

\subsection{Word Embeddings}

Using pre-trained word embeddings has become a standard feature of NLP tasks including NER. ~\citep{collobert:11} proposed a neural network architecture to construct word-embeddings that forms the primary method to get vector representation of words for training deep learning NLP models for NER. Word embeddings was pioneered by ~\citep{mikolov:13a} who introduced the continuous bag-of-words and the skip-gram models to build highly granular vector representation of words. Glove by ~\citep{pennington:14} is another famous word embedding method which is based on word co-occurences. The frequency matrix is factorised into a lower dimension by minimizing the reconstruction loss after normalisation and smoothing the matrix. The ~\citep{mikolov:13a} method to create word embeddings became widely adopted because this vector representation showed compositionality. Compositionality corresponds to the property of linear semantic inference eg. ‘Paris’-’France’ + ‘Italy’=’Rome’. 

\begin{figure}[h]
\includegraphics[scale=0.80]{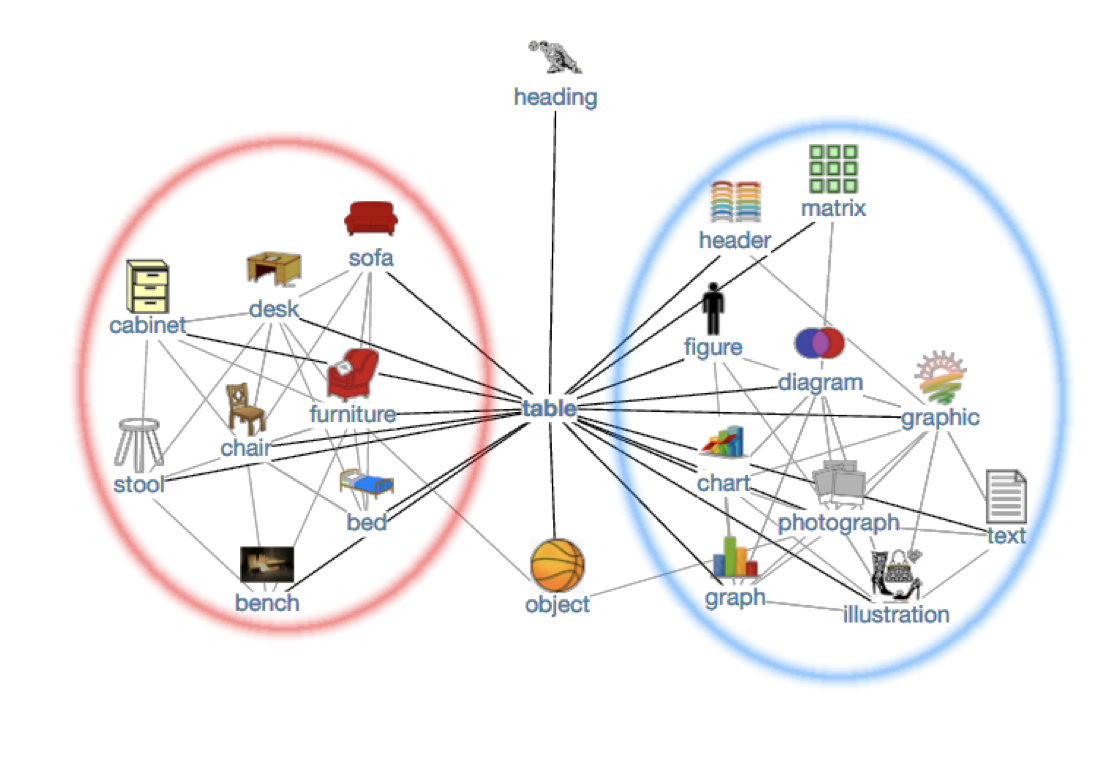}
\caption{Diagram of the graph obtained from the vector representation of furniture and data sense clusters related to the vector representation of the word 'table' (eg. ego network of 'table') ~\citep{pelevina:17}}
\centering
\end{figure}

The CBOW and the continuous skip-gram are both log-linear language models but they differ in a fundamental manner. The CBOW predicts the target word given its context. However the continuous skip gram model predicts words before and after a target word within a given window. The window of neighbouring words that is used as context for the vector representation is a hyperparameter that needs to be optimised. Increasing the window increases the accuracy of the language model but also increases the computational complexity of factoring in distant words in the window. One of the main advantages of the new log-linear models proposed by ~\citep{mikolov:13a} is that they eliminate the hidden layer in feed-forward neural net language models thus reducing computational bottleneck of the language model ie. an optimized single-machine implementation can train on more than 100 billion words in one day.
~\citep{mikolov:13b} further extended the original Skip-gram model for faster training and better quality word representation. They introduced a simple sub-sampling method to speed up the training process as well as ensure more accurate representation of less frequent words. The sub-sampling scheme is  to discard each word $w_i$ in the training set with probability computed by the formula $P(w_i) = 1- sqrt(/frac{t}{f(w_i)})$

Where $f(w_i)$ is the frequency of the word $w_i$ and t is the chosen frequency threshold (typically $10^-5$) above which words are sub-sampled significantly. Moreover, they replaced the computationally inefficient softmax method in the output layer with 2 alternatives :1) Hierarchical Softmax ( approximation of the full softmax) and 2) Noise Contrastive Estimation (NCE). Negative Sampling or NCE was introduced by ~\citep{gutmann:12} and applied to language modeling by ~\citep{mnih:12}.

There were caveats to extending the word embeddings to phrase embeddings ie. the football club ‘Manchester City’ is not the same in meaning  as the combination of the words ‘Manchester’ and ‘City’. Mikolov treated phrases as individual tokens in training the log-linear language model. Other methods ~\citep{johnson:15} have obtained the n-gram representation from unlabelled data.
Another issue with fixed word embeddings is that it does not account for polysemy ie a single word can be represented by 2 different vectors in 2 different contexts. Traditional word embedding methods such as Word2Vec and Glove consider all the sentences where a word is present in order to create a global vector representation of that word. However, a word can have completely different senses or meanings in the contexts. For example, let's consider these two sentences - 1) “The enjoyed reading the novel over the weekend” 2) “Scientists have discovered a novel method to treat cancer”. The word senses of ‘novel ’are different in these two sentences depending on its context. Traditional word embedding methods like word2vec and glove provide the same representation of ‘novel’ in both the contexts. ~\citep{upadhyay:17} used multilingual data to add another dimension to multi-sense word embeddings. For example, the English word bank, when translated to French provides two different words: banc and banque representing financial and geographical meanings, respectively. 

\subsection{Character Embeddings}

Character-level embeddings find usage in NER to capture the morphological features across languages. Better results on morphologically rich languages are reported in certain NLP tasks. ~\citep{santos:15} applied character-level representations, along with word embeddings for NER, achieving state-of-the-art results in Portuguese and Spanish corpora. ~\citep{kim:16} showed positive results on building a neural language model using only character embeddings. ~\citep{ma:16a} exploited several embeddings, including character trigrams, to incorporate prototypical and hierarchical information for learning pre-trained label embeddings in the context of NER. Chinese is another morphologically rich language in which character embeddings have shown better performance compared to word embeddings for deep learning sequence labelling models ~\citep{zheng:13}. 

\begin{figure}[h]
\includegraphics[scale=0.60]{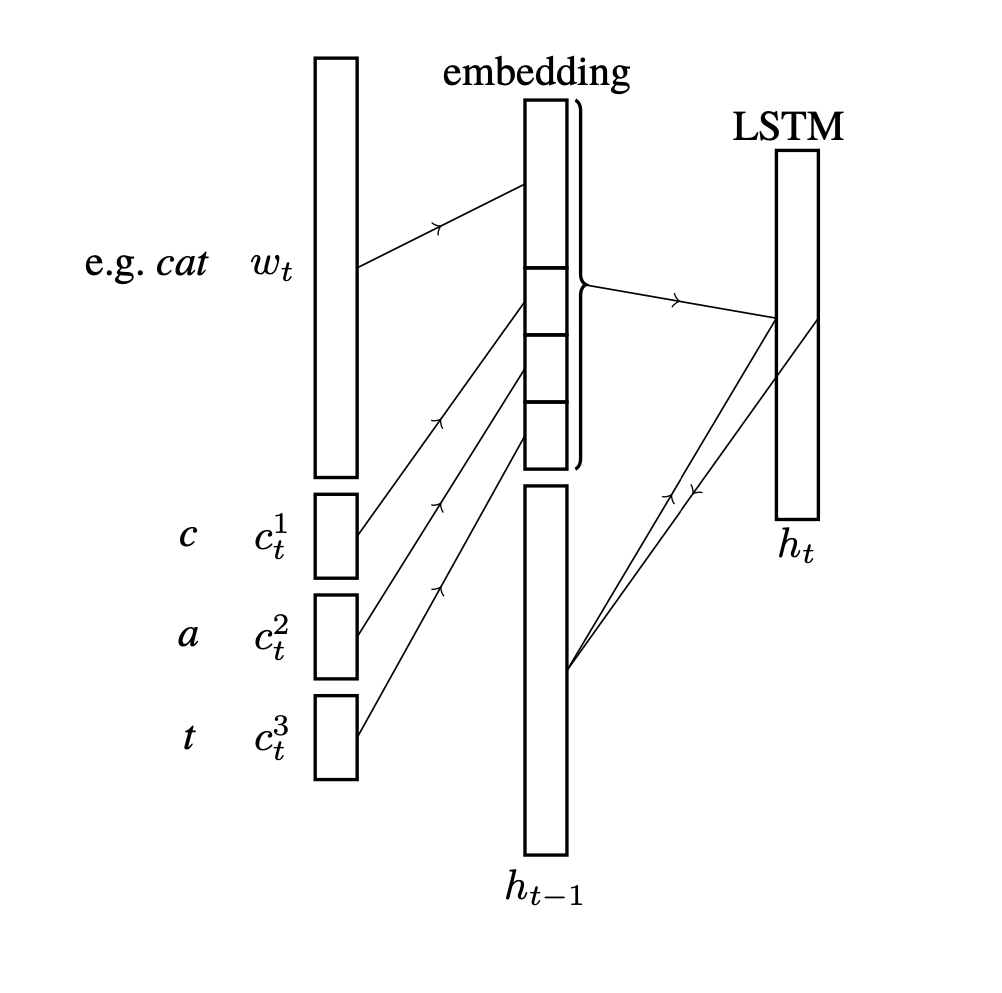}
\caption{Merging of the word and character embeddings of the word 'cat' in a LSTM LM (eg. language model) ~\citep{verwimp:17}}
\centering
\end{figure}

Word embeddings do not convey the semantic meaning and other internal information that character embeddings provide. Hence character embeddings are able to deduce the meaning of unknown words by mapping the meaning of such words to that of the compositional characters or sub-words. Therefore character embeddings solve the problem of identifying out-of-vocabulary (OOV) words eg. words that are not present in the input corpus for tasks like part-of-speech tagging and language modeling ~\citep{ling:15} or dependency parsing ~\citep{ballesteros:15}. Character embeddings provide a feasible method to represent word types. Character embeddings have found widespread use as they avoid the additional dimensions that are introduced to solve the OOV problem by strictly using word representations. ~\cite{chen:15} shows that introducing character embeddings to word embeddings in the Chinese language results in more informed word representation eg. better results in word relatedness and analogical reasoning tasks. Character embeddings also have the advantage of becoming relevant to the task, language and domain at hand ~\citep{vosoughi:16, saxe:17}. ~\cite{bojanowski:17} tried to improve the popular skip-gram method by representing words as bag-of-character n-grams. Therefore their work solved the limitations of word embeddings with the efficient skip-gram model. Their fastText library ~\citep{joulin:16, joulin:17} produces text classification models within 100kB without compromising on accuracy or speed.
The typical architecture involved in deriving word embeddings from characters initialises a character lookup table with random representation of each character. The character embeddings of every character in the sequence then passes through forward and backward LSTMs. The character representations from the biLSTM then concatenate with the word representation from a word lookup table. The forward LSTM represents the suffix of the word while the backward LSTM represents the prefix of the word. LSTM networks model long-range dependencies in temporal sequences better than traditional RNNs ~\citep{sak:14}. However, LSTM networks do not capture the semantic difference between the prefix and the suffix of a word that biLSTMs do. Convolutional neural networks have also been proposed to discover position invariant features of characters from words ~\citep{zhang:15 kim:16}. Efficient word selection further tackles the issue of non-compositional words or unknown words or ambiguous characters. 

Apart from character embeddings, different approaches have been proposed for OOV handling. ~\citep{herbelot:17} provided an essential component of incremental concept learning by initializing the unknown words as the greedy combination of the context words and refining these words with a high learning rate. However, their approach is yet to be tested on typical NLP tasks where the processing of context requires modulation. ~\cite{pinter:17} provided a character-based model that does not require re-training of character embeddings from the original corpus thus saving processing time. This allowed them to learn a compositional mapping form character to word embedding, thus efficiently tackling the OOV problem.

Despite the ever growing popularity of distributional vectors, there are also limitations of these vectors. For example, ~\citep{lucy:17} has recently tried to evaluate how well the word vectors predict the perceptual and conceptual features of distinct concepts, using a semantic norms dataset that has been created by human participants as a benchmark. The authors have discovered severe limitations of distributional models in making the fundamental understanding of the concepts behind the words. A possible direction for mitigating these deficiencies will be grounded learning ~\citep{niekum:13}.

\subsection{Model Specification}

\subsubsection{Convolutional Neural Networks}

CNNs found use in extracting feature vector representations from words ~\citep{collobert:11} using a lookup table learned by backpropagation. CNNs therefore seemed to be the natural choice in extracting higher order features from individual words in an input sequence of variable length. There are 2 approaches : (1) window approach and (2) (convolutional) sentence approach. The window approach assumes that the tag assigned to an individual word depends on its context ( ie. words occurring before and after the given word). The window approach is more suitable for sequence tagging tasks like NER. 

\begin{equation}
\begin{split}
    f^l_\theta = \langle LT_W ([w]^T_l) \rangle  ^{d_{win}}_t  = \begin{pmatrix} \langle W \rangle ^l _{[w]_{t-d_{win}/2}} \\ 
. \\
. \\
. \\
\langle W \rangle ^l _{[w]_{t }  } \\
. \\
. \\
. \\
\langle W \rangle ^l _{[w]_{t+d_{win}/2}} \end{pmatrix}
\end{split}
\end{equation}

\begin{equation}
    \begin{split}
        f^l_\theta = W^l f^{l-1}_\theta + b^l     
    \end{split}
\end{equation}

Here $W^l \in R^{n^l_h \times n^{l-1}_h}$ and $b^l \in R^{n^l_h}$ are trained using backpropagation. Here $n^l_h$ is a hyper-parameter that indicates the number  of hidden units in the $l^{th}$ layer. 
The fixed size vector input  can be passed through multiple linear transformation layers as shown above. 
In order to capture the higher-order features from the input sequence, there are layers of ‘’hard” hyperbolic tangent functions. The “hard” hyperbolic tangent is computationally efficient compared to the exact hyperbolic tangent function while preventing overfitting. 
One caveat of the window feature is that the context of the words at the beginning and the end of a sentence are not well-defined. Hence there are padding words half the size of the window at both the beginning and the end of input sentences akin to the start and stop indicators in sequence models.

\begin{equation}
    \begin{split}
        [ f^l_\theta ]_i = HardTanh( [  f^{l-1}_\theta ]_i ) 
    \end{split}
\end{equation}

\begin{equation}
    \begin{split}
        Hardtanh(x) = \begin{cases} -1 & if ~ x < -1 \\
x & if ~ -1\leq x\leq 1 \\
1 & if ~ x > -1 \end{cases}
    \end{split}
\end{equation}

The use of CNNs for sentence modeling traces back to ~\citep{collobert:08}. This work used multi-task learning to output multiple predictions for NLP tasks such as POS tags, chunks, named-entity tags, semantic roles, semantically-similar words and a language model. A look-up table was used to transform each word into a vector of user-defined dimensions. Thus, an input sequence $/{s_1 ~, s_2 ~, ...s_n ~/}$ of n words was transformed into a series of vectors $/{ws_1 ~ , ws_2 ~, ...ws_n ~/}$ by applying the look-up table to each of its words.

\begin{figure}[h]
\includegraphics[scale=0.60]{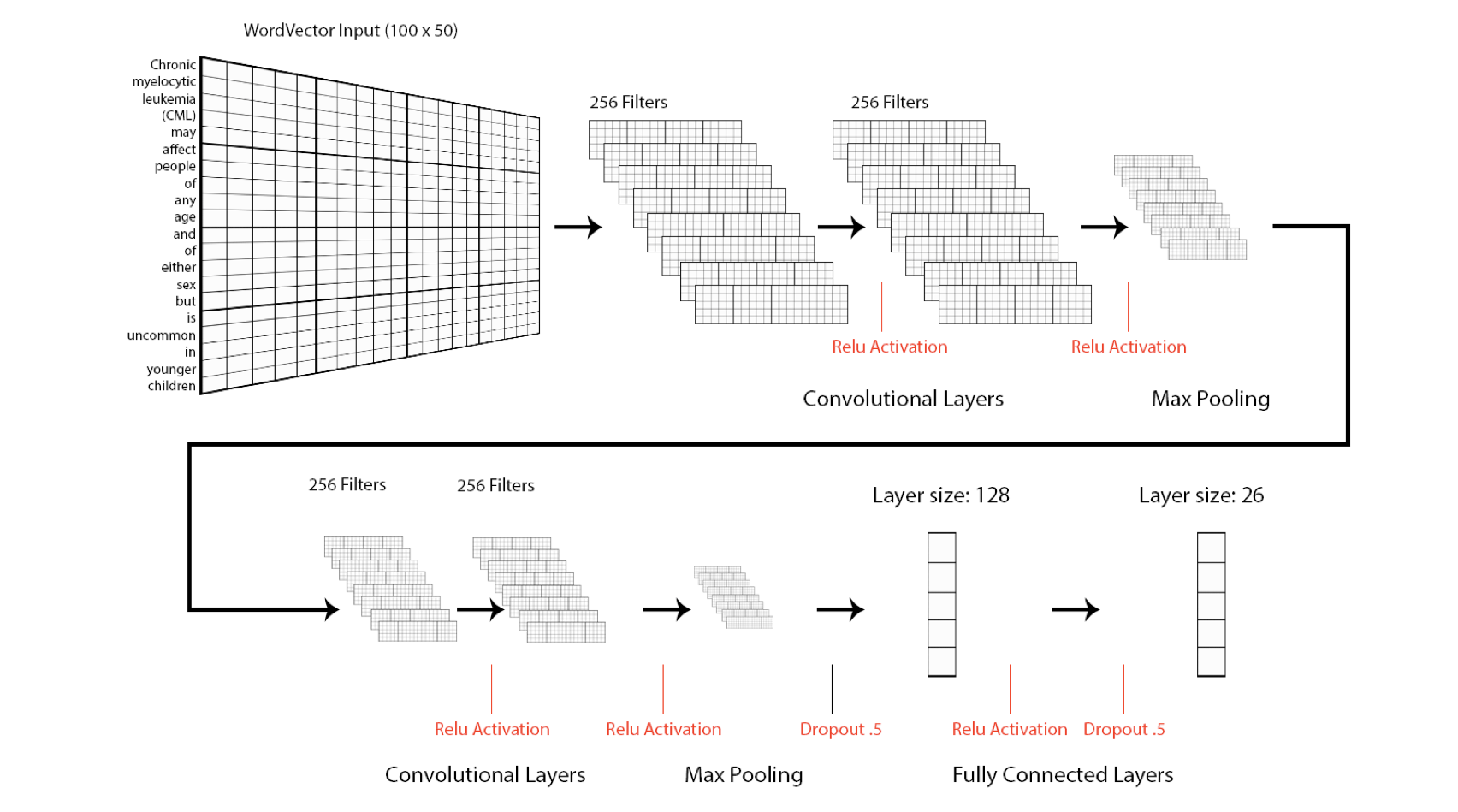}
\caption{Example of a CNN architecture for medical text classification ~\citep{hughes:17}}
\centering
\end{figure}

This can be thought of as a primitive word embedding method whose weights were learned in the training of the network. In ~\citep{collobert:11}, Collobert et al extended their work to propose a general CNN-based framework to solve a plethora of NLP tasks. Both these works triggered a huge popularization of CNNs amongst NLP researchers. Given that CNNs had already shown their mettle for computer vision tasks, it was easier for people to believe in their performance.

CNNs have the ability to extract salient n-gram features from the input sentence to create an informative latent semantic representation of the sentence for downstream tasks. This application was pioneered by ~\citep{collobert:11, kalchbrenner:14, kim:14} which led to a huge proliferation of CNN-based networks in the succeeding literature. Below, we describe the working of a simple CNN-based sentence modeling network:

\subsubsection{Recurrent Neural Network}

Recurrent Neural Network ~\citep{elman:90} is another form of deep neural network architecture. In general, CNNs are used to represent position-invariant functions (eg bag of words) while RNNs represent a sequential architecture (eg. sentences, paragraphs etc.). It seems obvious that RNNs would be more suitable for sequence modelling tasks like NER compared to the hierarchical CNNs. While we saw that the window approach helps to tackle sequence inputs in CNN, there is no way to model dependencies on context beyond that which is fixed in cross-validation. RNNs help to capture dependencies on words or sentences beyond the fixed range of CNNs. The simple RNN has no gating mechanism. RNN is a network that is unfolded across time thus providing a spatial representation of memory. For a given input, the RNN calculates a hidden state as follows:

\begin{equation}
    \begin{split}
        [s_{\theta,l}^t]_i = g_{\theta, l} ([f_{\theta, l}^t]_i )  
    \end{split}
\end{equation}

\begin{equation}
    \begin{split}
         [f_{\theta, l}^t]_i = W^l [x_l^t]_i + [s_{\theta, l}^{t-1}]_i
    \end{split}
\end{equation}

Here $ [s^t_{\theta, l}]_i $ is the hidden state at time t and layer l for the $ i_th$ unit of the input sequence. $ g_{\theta, l} $ is a nonlinear function ( eg. tanh etc.) at the layer which takes as input $ f_{\theta, l}^t $. $ [x_l^t]_i $ is the ith unit ( word, sentence etc.) of the input sequence at time t and $ W^l \in R^{d_{s^t_{\theta,l}} \times d_{x^t_l}} $ are the weights that are learned through training. 
The hidden state in a simple RNN can be considered its memory component. However, simple RNNs suffer from the vanishing gradient problem that makes it difficult to learn the weights in the previous time steps using backpropagation. Therefore simple RNNs are augmented with a gating mechanism to overcome the convergence problem. RNN variants with gating mechanism that are most popular in NER are long-short term memory (LSTM) and Gated Recurrent Units (GRUs). 

\subsubsection{Long Short Term Memory}

The novelty of LSTM ~\citep{hochreiter:97} is that it is able to bridge long time intervals (to quickly learn the slowly changing weights many time steps back eg. long term memory) as well as preserve the recent inputs ( eg. short term memory). Moreover, the LSTM architecture ensures constant re-weighting thus avoiding the explosion or vanishing of error flow through the hidden states. 

\begin{equation}
    i_t = \sigma ( x_t U ^i + h_{t-1} W^i + b_i )
\end{equation}
\begin{equation}
    f_t = \sigma (x_t U^f + h_{t-1} W^f + b_f )
\end{equation}
\begin{equation}
    o_t = \sigma (x_t U^o + h_{t-1} W^o + b_o )
\end{equation}
\begin{equation}
    q_t = tanh (x_t U^q + h_{t-1} W^q + b_q )
\end{equation}
\begin{equation}
    p_t = f_t \times p_{t-1} + i_t \times q_t
\end{equation}
\begin{equation}
    h_t = o_t \times tanh (p_t )
\end{equation}

LSTM has three gates: input gate $i_t$, forget gate $f_t$ and output gate $o_t$ that are outputs of the sigmoid function over the input $x_t$ and the preceding hidden state $h_{t-1}$. In order to generate the hidden state at current step t, it first generates a temporary variable $q_t$  by running the non-linear tanh function on the input $x_t$ and the previous hidden state $h_{t-1}$. The LSTM then calculates an updated history variable at time t $ p_t$ as a linear combination of the previous history state $p_{t-1}$ and the current temporary variable $q_t$ scaled by the current forget gate $f_t$ and the current input gate $i_t$ respectively. Finally LSTM runs the tanh on $p_t$ and scales it by the current output gate $o_t$ to get the updated hidden state $h_t$ . 

\begin{figure}[h]
\includegraphics[scale=0.50]{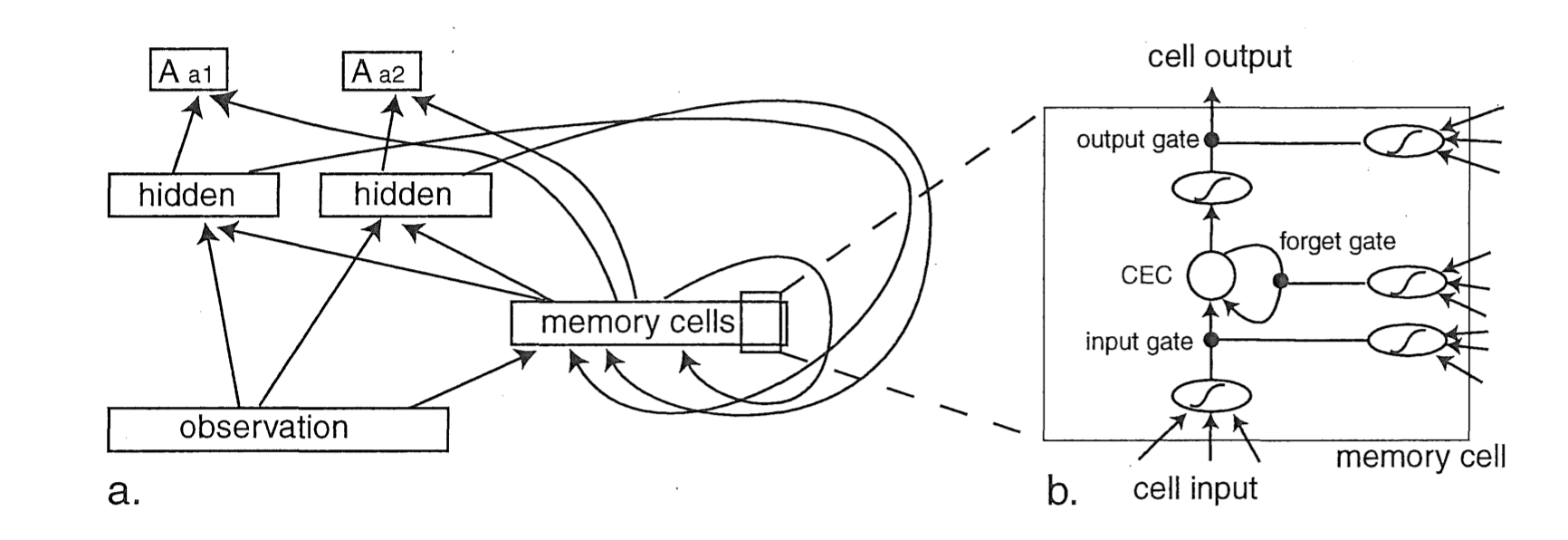}
\caption{a. General LSTM architecture b. One memory cell ~\citep{bakker:02}}
\centering
\end{figure}

While LSTM is good at approximating the dependency of the current unit of the sequence with previous units, it does not take into account the dependency of the current unit on units to its right in the sequence. ~\citep{lample:16} solved this problem by implementing a bi-directional LSTM ~\citep{graves:05}. In other words, there are 2 independent LSTMs that are trained separately using the segment of the input sequence to the left and the right of the target word. The model gets the full representation of the target word by concatenating the left and right context representations of the forward and the backward LSTM respectively ie the hidden state $ h_t = [ \overrightarrow h_t ; \overleftarrow h_t] $.

\subsubsection{Gated Recurrent Unit}

Gated Recurrent Unit ~\citep{cho:14} is a more recent and less complex variant of the RNN compared to the LSTM. 

\begin{figure}[h]
\includegraphics[scale=0.70]{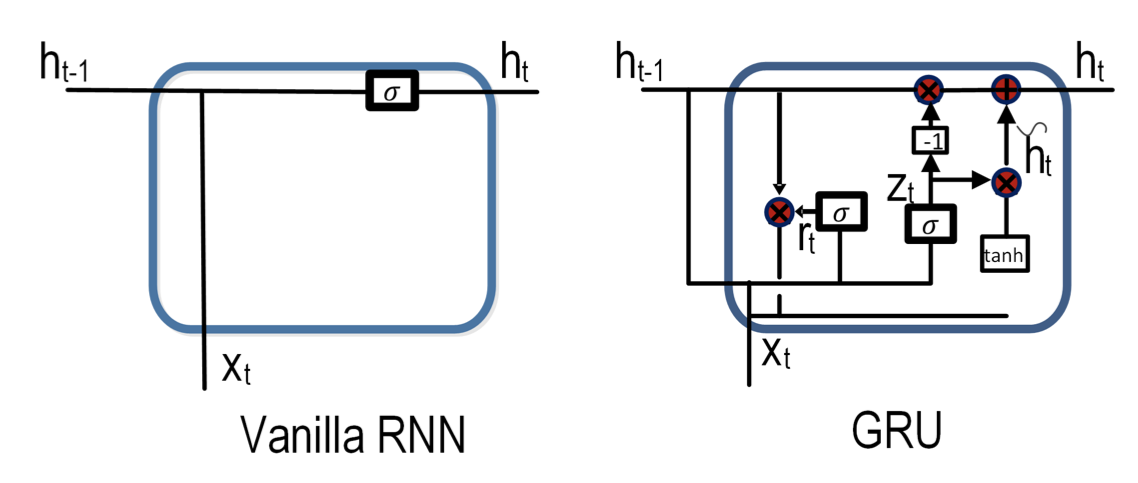}
\caption{Vanilla RNN and GRU architecture (single unit shown for simplicity) with a reset gate r (adjusts the incorporation of new input with the previous memory), an update gate z (controls the preservation of the previous memory), current hidden output $h_t$ and previous hidden output $h_{t-1}$ ~\citep{zhao:17}}
\centering
\end{figure}

The GRU is similar to the LSTM as it modulates the error flow thus avoiding vanishing gradients ~\citep{bengio:94}. However, GRU has many crucial differences with the LSTM. GRUs do not have separate memory cells like LSTMs. Therefore GRUs lack the controlled exposure to the memory content (ie the output gate of LSTMs). Unlike LSTMs, GRUs expose the full content without any control. Moreover, GRUs control the information flow from the previous activation to the current, candidate activation. On the other hand, LSTMs compute the latest memory content without any control over the historical information flow. The GRU works using the following equations:

\begin{equation}
    h^j_t = ( 1 - z^j_t) h^j_{t-1} + z^j_t \bar h^j_t 
\end{equation}
\begin{equation}
    z^j_t = \sigma ( W_z x_t + U_z h_{t-1} )
\end{equation}
\begin{equation}
    \bar h^j_t = tanh (W x_t + U (r_t \odot h_{t-1}))
\end{equation}
\begin{equation}
    r^j_t = \sigma ( W_r x_t + U_r h_{t-1} )
\end{equation}

$h^j_t$ is the activation of the GRU at level j at time t. $ \bar h^j_t $ is the candidate activation of the GRU ~\citep{bahdanau:14} at level j at time t. The update gate $z^j_t$ decides how much the unit updates its activation. The reset gate $r^j_t$ is computed similar to the update gate.

\subsubsection{Results}

Tables \ref{table:1} and \ref{table:2} show the F1 scores of state of the art neural network models in chronological order for NER on the dev as well as test data set from CoNLL-2003 shared task and the OntoNotes 5.0 NER task (English). Tables \ref{table:3} \ref{table:4} show the F1 scores of the state of the art models for the CoNLL 2002 Spanish and Dutch NER shared tasks. 

\begin{table}[h!]
\centering
\begin{tabular}{ |p{3cm}|p{1cm}|p{1cm}|p{2.75cm}|p{1cm}|p{1cm}|p{2.75cm}|  }
 \hline
 \multirow{2}{*}{\textbf{Model}} & \multicolumn{3}{c|}{\textbf{CoNLL-2003}} & \multicolumn{3}{c|}{\textbf{OntoNotes 5.0}} \\
 \cline{2-7} & \textbf{Prec.} & \textbf{Recall} & \textbf{F1} & \textbf{Prec.} & \textbf{Recall} & \textbf{F1} \\
 \hline
  ~\cite{collobert:11} & - & - & 88.67 & - & - & - \\
\hline
  ~\cite{collobert:11}+lexicon & - & - & 89.59 & - & - & - \\
\hline
~\cite{huang:15} & - & - & 84.26 & - & - & - \\
\hline  
   ~\cite{huang:15}+Senna + gazetteer & - & - & 90.10 & - & - & - \\
\hline   
~\cite{chiu:15} & 83.48 & 83.28 & 83.28($\pm$ 0.20) & 82.58 & 82.49 & 82.53($\pm$ 0.40) \\
\hline
 ~\cite{chiu:15} + emb & 90.75 & 91.08 & 90.91($\pm$ 0.20) & 85.99 & 86.36 & 86.17($\pm$ 0.22) \\
\hline
 ~\cite{chiu:15} + emb + lex & 91.39 & 91.85 & 91.62($\pm$ 0.33) & 86.04 & 86.53 & 86.28($\pm$ 0.26) \\
\hline
 ~\cite{ma:16b} & - & - & 91.37 & - & - & - \\
\hline
   ~\cite{lample:16} (Bi-LSTM) & - & - & 89.15 & - & - & - \\
\hline   
   ~\cite{lample:16} (Bi-LSTM-CRF) & - & - & 90.94 & - & - & - \\
\hline   
   ~\cite{hu:16} & - & - & 91.18 & - & - & - \\   
\hline
\end{tabular}
\caption{Table to test captions and labels}
\label{table:1}
\end{table}

There are 22 model results in Tables \ref{table:1} and \ref{table:2} while there are 8 model results in Tables \ref{table:3} and \ref{table:4}. For the purpose of presentation, the neural models are divided into 2 tables with 11 results in each table. Tables \ref{table:3} and \ref{table:4} show the F1 scores of the linear and the log-linear models ( ie. does not involve neural networks) for NER. The latest NER model that does not have any form of neural network in its architecture was produced in 2015. All the NER models that have produced state of the art results since 2015 use neural network architecture. Therefore, it is unlikely that linear methods will be able to compete with the results produced by deep learning models for NER and other sequence tagging tasks. 

\begin{table}[h!]
\centering
\begin{tabular}{ |p{3cm}|p{1cm}|p{1cm}|p{2.75cm}|p{1cm}|p{1cm}|p{2.75cm}|  }
 \hline
 \multirow{2}{*}{\textbf{Model}} & \multicolumn{3}{c|}{\textbf{CoNLL-2003}} & \multicolumn{3}{c|}{\textbf{OntoNotes 5.0}} \\
 \cline{2-7} & \textbf{Prec.} & \textbf{Recall} & \textbf{F1} & \textbf{Prec.} & \textbf{Recall} & \textbf{F1} \\
 \hline
 ~\cite{yang:16} - no word embeddings & - & - & 77.20 & - & - & - \\ 
\hline
 ~\cite{yang:16} - no char GRU & - & - & 88.00 & - & - & - \\   
\hline
 ~\cite{yang:16} - no gazetteer & - & - & 90.96 & - & - & - \\   
\hline
 ~\cite{yang:16} & - & - & 91.20 & - & - & - \\   
\hline
~\cite{rei:17} & - & - & 87.38($\pm$0.36) & - & - & - \\   
\hline
   ~\cite{strubell:17} & - & - & 90.54 & - & - & - \\   
\hline
   ~\cite{yang:17} & - & - & 91.20 & - & - & - \\   
\hline
   ~\cite{yang:17} + CoNLL 2000/PTB-POS & - & - & 91.26 & - & - & - \\   
\hline
   ~\cite{peters:17}(word embedding) & - & - & 90.87($\pm$0.13) & - & - & - \\ 
\hline
   ~\cite{peters:17}(LM embedding) & - & - & 90.79($\pm$0.15) & - & - & - \\ 
\hline
   ~\cite{peters:17} + 1B Word dataset & - & - & 91.62($\pm$0.23) & - & - & - \\    
\hline
   ~\cite{peters:17} + 1B Word dataset + 4096-8192-1024 & - & - & 91.93($\pm$ 0.19) & - & - & - \\   
\hline
\end{tabular}
\caption{Table to test captions and labels}
\label{table:2}
\end{table}

It is also clear from Table \ref{table:1} that the use of character-level and word-level knowledge ( eg. gazetteers, benchmark corpus like ~\citep{chelba:13}, LM ie language model) benefits neural models. Specifically word embeddings ~\citep{mikolov:13a, pennington:14} and sub-word or character embeddings are used as additional features to the input text corpus for co-training.  

\begin{table}[h!]
\centering
\begin{tabular}{ |p{3cm}|p{1cm}|p{1cm}|p{1cm}|p{1cm}|p{1cm}|p{1cm}|  }
\hline
 \multirow{2}{*}{\textbf{Model}} & \multicolumn{3}{c|}{\textbf{CoNLL-2003}} & \multicolumn{3}{c|}{\textbf{OntoNotes 5.0}} \\
 \cline{2-7} & \textbf{Prec.} & \textbf{Recall} & \textbf{F1} & \textbf{Prec.} & \textbf{Recall} & \textbf{F1} \\
 \hline
~\cite{chieu:02} & - & - & 88.31 & - & - & - \\
\hline
~\cite{florian:03} & - & - & 88.76 & - & - & - \\
\hline
~\cite{ando:05}(dev) & - & - & 93.15 & - & - & - \\
\hline
~\cite{ando:05}(test) & - & - & 89.31 & - & - & - \\
\hline
~\cite{finkel:05} & 95.10 & 78.30 & 85.90 & - & - & - \\
\hline
~\cite{suzuki:08}(dev) & - & - & 94.48 & - & - & - \\
\hline
~\cite{suzuki:08}(test) & - & - & 89.92 & - & - & - \\
\hline
\end{tabular}
\caption{Table to test captions and labels}
\label{table:3}
\end{table}

Moreover, most of the recent LSTM models (eg. LSTM-CRF ~\citep{huang:15, lample:16}, LSTM-CNNs ~\citep{chiu:15}) utilize syntax (eg. character type, capitalization), lexicon, context, POS features as well as preprocessing to suit the NER task. ~\cite{Lou:15} achieved a high F! score (91.2) by using hand-crafted features like word, character, POS, chunk and stemmed n-grams, Brown and WordNet clusters as well as dictionaries from external knowledge bases like Wikipedia and Freebase. 

\begin{table}[h!]
\centering
\begin{tabular}{ |p{3cm}|p{1cm}|p{1cm}|p{1cm}|p{1cm}|p{1cm}|p{1cm}|  }
\hline
 \multirow{2}{*}{\textbf{Model}} & \multicolumn{3}{c|}{\textbf{CoNLL-2003}} & \multicolumn{3}{c|}{\textbf{OntoNotes 5.0}} \\
 \cline{2-7} & \textbf{Prec.} & \textbf{Recall} & \textbf{F1} & \textbf{Prec.} & \textbf{Recall} & \textbf{F1} \\
 \hline
  ~\cite{ratinov:09} & 91.20 & 90.50 & 90.80 & 82.00 & 84.95 & 83.45 \\
 \hline
 ~\cite{lin:09} & - & - & 90.90 & - & - & - \\
 \hline
 ~\cite{finkel:09} & - & - & - & 84.04 & 80.86 & 82.42 \\
 \hline
 ~\cite{suzuki:11} & - & - & 91.02 & - & - & - \\
 \hline
 ~\cite{sil:13} & 86.80 & 89.50 & 88.20 & - & - & - \\
 \hline
 ~\cite{passos:14} & - & - & 90.90 & - & - & 82.24 \\
\hline
  ~\cite{durrett;14} & - & - & - & 85.22 & 82.89 & 84.04 \\
\hline
~\cite{luo:15} (NER features)  & 90.00 & 89.90 & 89.90 & - & - & - \\
  \hline
  ~\cite{luo:15} (NER + EL features)  & 91.50 & 91.40 & 91.20 & - & - & - \\
  \hline
 \end{tabular}
\caption{Table to test captions and labels}
\label{table:4}
\end{table}

Such additional information also finds use as pre-trained embeddings to downstream models. Although the pre-trained models are not task-specific, they improve the performance of NER models. For example, ~\cite{chiu:16} reports a significant improvement (Wilcoxon ranked sum test, p < 0.001) when they train their model with word embeddings ~\citep{collobert:11} compared to random embeddings regardless of the additional features used for training. Co-training strategies are more favorable compared to pre-training strategies since the former requires a less complex neural network compared to the latter. Therefore co-training is more efficient because it does not require any additional corpus and hence the training time is also less. 

\begin{table}[h]
\centering
\begin{tabular}{ |p{3cm}|p{2cm}|p{2cm}|  }
\hline
 \multirow{2}{*}{\textbf{Model}} & \textbf{CoNLL 2002 Dutch NER} & \textbf{CoNLL 2002 Spanish NER} \\
 \cline{2-3} & \textbf{F1 (\%)} & \textbf{F1 (\%)} \\
 \hline
 ~\cite{carreras:02} & 77.05 & 81.39  \\
\hline
 ~\cite{nothman:13} & 78.60 & -  \\
\hline
 ~\cite{gillick:15} & 82.84 & 82.95  \\
 \hline
 ~\cite{lample:16} & 81.74 & 85.75  \\
 \hline
  ~\cite{yang:16} - no word embeddings & 67.36 & 73.34  \\
 \hline
  ~\cite{yang:16} - no char GRU & 77.76 & 83.03  \\
 \hline
  ~\cite{yang:16}  & 85.00 & 84.69  \\
 \hline
  ~\cite{yang:16} + joint training & 85.19 & 85.77  \\
 \hline
\end{tabular}
\caption{Table to test captions and labels}
\label{table:5}
\end{table}

Efforts to increase NER F1 score by pre-training, co-training or joint training hss come at a cost. With respect to a generic neural network architecture, additional training layers significantly increases the complexity and the training time of the NER models making the use of these models difficult in practice. ~\cite{strubell:17} introduced the use of dilated convolutions that gave close to state of the art results while improving efficiency over existing models by processing a larger input window at a time to model context in a parallelized manner. Another aspect that is evident from Table \ref{table:1} is that CRF is very suitable as the output layer of neural network NER models.
 
% Acknowledgements should go at the end, before appendices and references

\acks{We would like to acknowledge support for this project
from the Dr. Geraldo Filgueiras }


\newpage

\vskip 0.2in

\bibliography{paper}

\end{document}